\documentclass{article}

     \usepackage[final]{neurips_2021}

\usepackage[utf8]{inputenc} 
\usepackage[T1]{fontenc}    
\usepackage{hyperref}       
\usepackage{url}            
\usepackage{booktabs}       
\usepackage{amsfonts}       
\usepackage{nicefrac}       
\usepackage{microtype}      
\usepackage{xcolor}         
\usepackage{graphicx}
\usepackage{caption}
\usepackage{subcaption}
\usepackage{algorithm}
\usepackage{algorithmic}
\usepackage{array}
\usepackage{soul}

\title{An Automated System for Detecting Visual Damages of Wind Turbine Blades}

\author{%
   Linh Nguyen \\
   SkySpecs Inc. \\
  \small{\texttt{linh.nguyen@skyspecs.com}}
  \And
 Akshay Iyer\\ 
  SkySpecs Inc.\\
  \small{\texttt{akshay.iyer@skyspecs.com}}
   \And
   Shweta Khushu\\
   SkySpecs Inc.  \\
 \small{\texttt{shweta.khushu@skyspecs.com}}
}

\begin{document}

\maketitle

\begin{abstract}

Wind energy's ability to compete with fossil fuels on a market level depends on lowering wind's
high operational costs. Since damages on wind turbine blades are the leading cause for these 
operational problems, identifying blade damages is critical. However, recent works in 
visual identification of blade damages are still experimental and focus on optimizing 
the traditional machine learning metrics such as IoU.
In this paper, we argue that pushing models to production long before achieving the "optimal" model performance
can still generate real value for this use case. 
We discuss the performance of our damage's suggestion model in production and how this system works in coordination with humans as part 
of a commercialized product and how it can contribute towards lowering wind energy's operational costs. 

\end{abstract}

\section{Introduction}
\label{intro}

In order to replace fossil fuels, renewable energy sources such as wind need to achieve a levelized cost  
competitive to that of the former, which is primarily responsible for CO$_2$ emissions \cite{iea2021}. 
One of the major hurdles to wind energy is the high operational cost, which accounts for 
about 25\% of the cost of energy compared to around 10\% for natural gas \cite{irena}. During the operations of 
wind turbines, damages on blades is the leading cause of turbine failures \cite{blade}. The benefits to
detecting these damages early are two fold: keeping the turbines less susceptible to downtime, and lowering
the costs of fixing these damages by not allowing damages to worsen over time. These benefits are key because ultimately, a wind turbine needs to be
operational to generate revenue.

The use of drones with cameras offers an attractive solution to finding damages on blades. Autonomous drones
can traverse the blades and take pictures along the way with minimal to no human intervention. To this end, we 
have built and commercialized a drone inspection system.  After each turbine inspection, the pictures are 
uploaded to a portal that then presents these pictures to a team of trained analysts to identify and 
categorize the damages.

\begin{figure}[ht]
     \centering
     \begin{subfigure}[t]{0.23\textwidth}
         \centering
         \includegraphics[width=\textwidth]{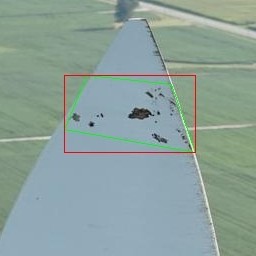}
         \caption{Ground truth polygon (green)  and bounding box (red)}
         \label{fig:ground_truth_bounding_box}
     \end{subfigure}
     \begin{subfigure}[t]{0.23\textwidth}
         \centering
         \includegraphics[width=\textwidth]{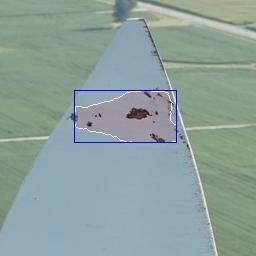}
         \caption{Damage predicted by our system: Prediction mask and minimum area rectangle around it (blue)}
         \label{fig:blue_marked_damage}
     \end{subfigure}
     \begin{subfigure}[t]{0.23\textwidth}
         \centering
         \includegraphics[width=\textwidth]{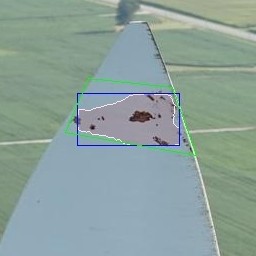}
         \caption{Agreement between analysts and our system}
         \label{fig:agreement_boxes}
     \end{subfigure}
     \begin{subfigure}[t]{0.23\textwidth}
         \includegraphics[width=\textwidth]{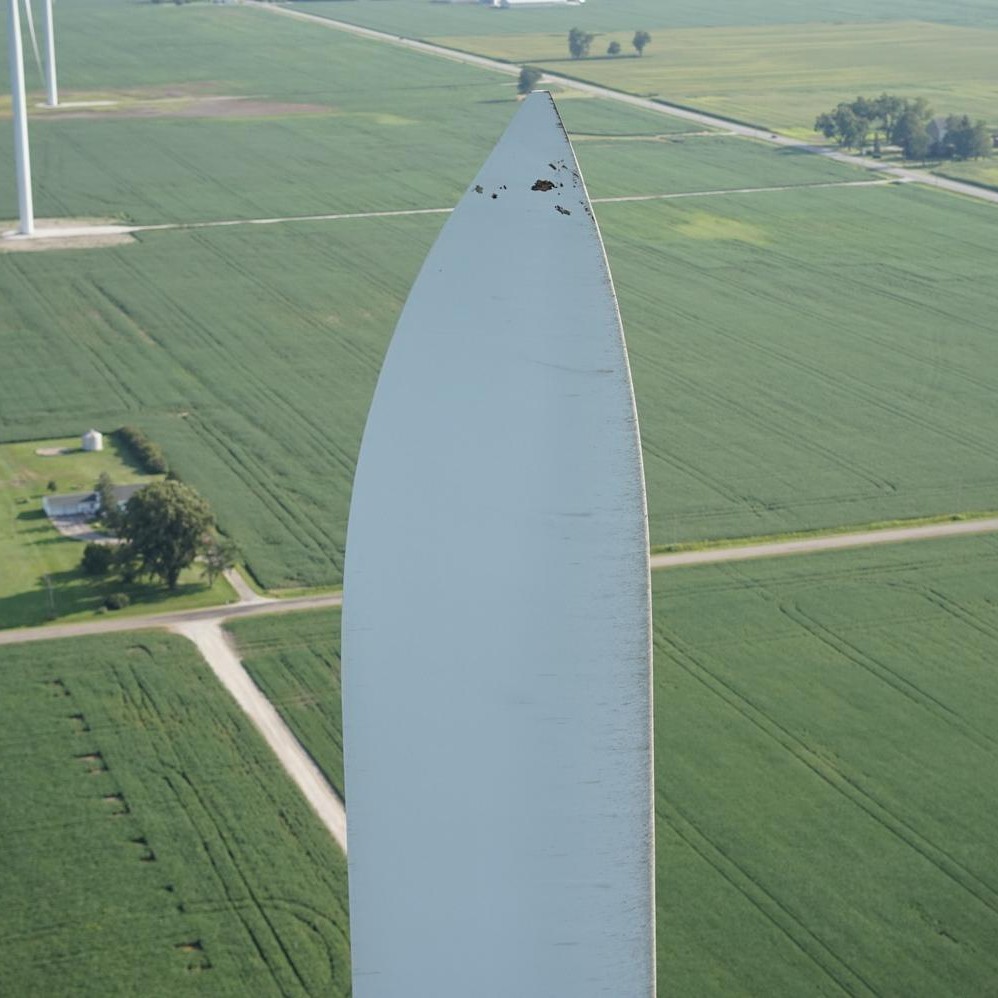}
         \caption{Original inspection picture. Note that the damage is at the tip of the blade}
         \label{fig:inspections_org}
     \end{subfigure}
\end{figure}

In light of lowering the cost of maintenance, analysts need to: 1. Miss fewer damages, and 2. Take less time
to mark damages. Traditionally, analysts find damages manually by drawing a polygon around the area of damage,
as shown in \ref{fig:ground_truth_bounding_box}. However, sometimes even experienced analysts miss damages.
The reason is that the blade is fairly large with respect to the damages, as shown in \ref{fig:inspections_org}. In addition, there are often multiple damages in one picture. Analysts can also miss
damages simply because of fatigue. We aim to help analysts increase accuracy and productivity through an automated damage suggestion system.

Recognizing the importance of this problem, a number of authors have proposed different approaches for damage detection of
wind turbines blades from drone images. 
 
A two-stage learning approach was proposed in \cite{icarm2017}. First, a feature extractor was trained, then a classifier to categorize damages on blades.
Another work attempted to classify damages on close-up pictures using different deep learning architectures \cite{CCC2020}. \cite{stokkeland14} conducted a wind
turbine inspection with a multicopter and used traditional computer vision algorithm to detect damages. \cite{dtu} proposed an automated suggestion system for damage detection 
that was based on Faster-RCNN. The system's mAP scores were compared to that of humans.

These works are critically valuable in discovering new architectures or methodologies to ML problems.
However, the real-world benefits often cannot be realized in the performances on test datasets or baked into a loss function. 
Neither the amount of time spent on annotating inspection pictures nor the chance of missing damages is tunable at the model level. Hence, our solution is to put the models in front 
of analysts quickly and to place more weight on iteratively optimizing 
\textit{how the model was used}, rather than focusing on achieving better results on a fixed
test dataset. We believe that this is fundamentally different from prior works in this area.

In this work, we describe an automated damage suggestion system, designed specifically to achieve these two goals.
We establish the following criteria, which better measures how successful a blade
damage detection model is, as such a system must be evaluated by the values driven \textit{for users}. These values are rarely manifested in the common process of tuning a machine learning (ML) model.

\begin{itemize}
    \item The model must achieve high \textit{damage recall}. That is, the model should be able to produce
        predictions that are close to the actual damage. The predictions and ground truths
        do not have to be an exact overlap, but do need to be greater than a certain
        IoU threshold. More than one prediction per actual damage is acceptable, as
        long as the actual damage is not missed. Note that we ignore pixel-level recall entirely.

    \item The integration of the model into the current workflow must not slow down analysts
        in annotating blade damages. We achieve this by presenting the predictions
        from the model as polygons analysts can simply choose from instead of having to
        draw from scratch. Moreover, we frequently ask for direct feedback whether this
        approach is efficient.
\end{itemize}

\section{Methods and Results}

\subsection{Inspection Data}

The onboard camera is the Sony UMC-R10C with a resolution of 5456x3632. These pictures are then uploaded to our portal so analysts can access and annotate them. These analysts
are trained by blade experts. To date, we have inspected over 90,000 wind turbines, both onshore and offshore, in more than 26 countries. 

\subsection{Network}
\label{blue}

The model is based on Mask R-CNN \cite{mask-rcnn}, which was implemented in PyTorch using 
a publicly available toolbox called detectron2 \cite{wu2019detectron2}.
For the purpose of training, we randomly selected more 
than 250,000 inspection pictures with over 370,000 annotations similar to that in Figure \ref{fig:inspections_org}. This dataset is splitted into train, test, validation sets at
80\%, 10\%, and 10\%, respectively. The model was trained on a \textit{p3.2xlarge} AWS EC2
instance with a NVIDIA Tesla V100.

Identifying damages is challenging because there are several types of damages that 
are visually very different
from each other. Moreover, shadows, stains, weather marks, and auxiliary components on the blades can give
a false appearance of damages. The distribution of damages is also highly uneven. Following our 
main insight that the model does not need to predict all of this information to be valuable to analysts,
we decided to treat all damages the same and the decision to classify these damages should be left to humans.

Since the original resolution was large, we down-sampled the images to 1500x998 for training. 
The ground truth masks were the polygons drawn by analysts and the bounding boxes were generated
by taking the max x and y coordinates from these polygons. The masks and bounding boxes are denoted
green and red, respectively, in Figure \ref{fig:ground_truth_bounding_box}. The problem 
formulated as detection followed by segmentation.

\subsection{The Quality Control (QC) process}

The predictions generated by the model are termed \textit{clues}. A clue is simply the tightest box (minimum area rectangle)
around the predicted masks, as illustrated in Figure \ref{fig:blue_marked_damage}. We present these
clues to analysts to choose from, and modify as they wish. Figure \ref{fig:agreement_boxes} 
demonstrates our key insight that the ground truths and predicted masks do not need to overlap
significantly to capture the damage. 

Previously, analysts would see the raw inspection pictures, find the damages, and manually
draw polygons around them. With our system, analysts simply review and approve the clues. This is 
a significantly easier process. The annotations undergo a second quality check by a human. Note that
the second-stage human quality check had already existed with the original process, thus is not an added task by using our system.

\begin{figure}[ht]
\vskip 0.2in
\begin{center}
\centerline{\includegraphics[width=\columnwidth]{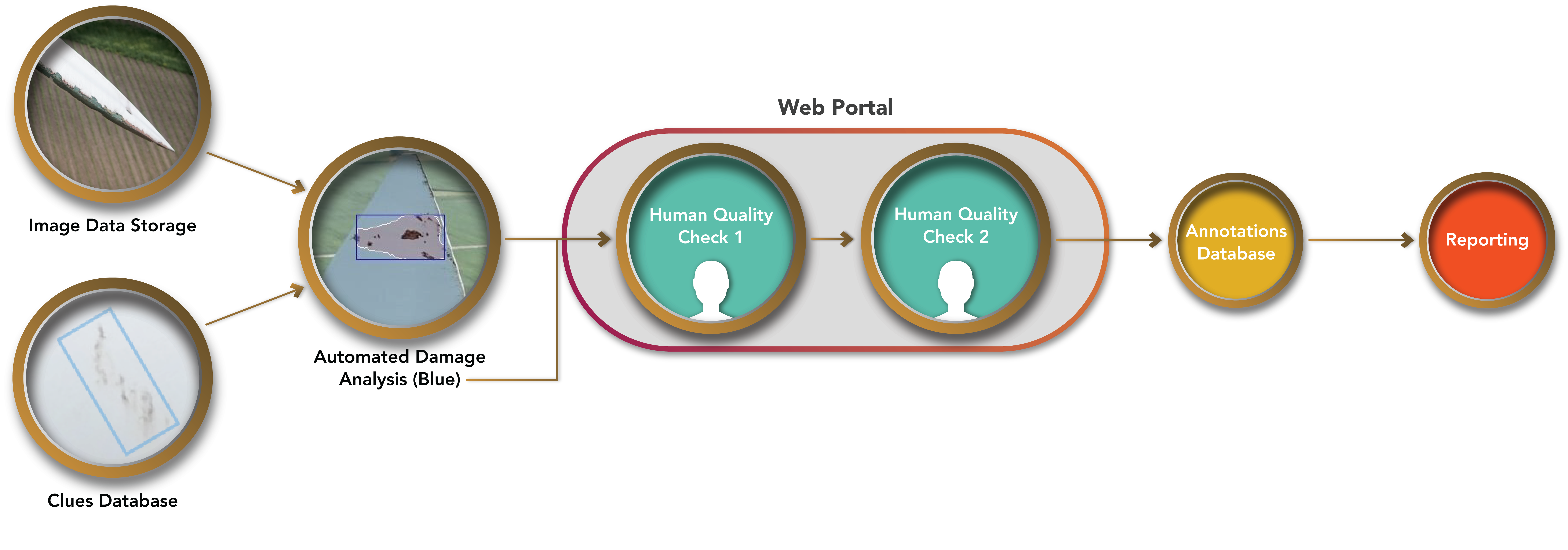}}
\caption{The new QC Process, with the new model in the loop}
\label{fig:qc-process}
\end{center}
\vskip -0.2in
\end{figure}

\section{Results}

Since it is much more important to have fewer missed damages, rather 
than \textit{detecting the exact damage}, we feel that the traditional approach for evaluating detection models 
is not sufficient. With this insight, we compute metrics to capture the 
precision and recall of \textit{damages} rather than pixels. 
All predictions are iteratively examined. 
A single true positive is an outcome when a ground truth polygon's IoU with one or many prediction polygons is greater than a pre-defined threshold.

On a test dataset of 18318 images, we achieved a damage recall of 94.4\% and damage precision of 40.48\%. 
Following our belief that the model needs to be tested on live inspections quickly, rather 
than continuing optimizing, we let the model go live in an A/B testing manner. We keep both the previous process 
(without) clues and the new process with clues for our analysts. 80\% of inspections were done with the former while the
rest were done with latter.

In order to monitor the production performance, we have built a dashboard that captures the metrics needed to measure 
the two evaluation criteria: recall and productivity. We measure productivity by comparing the normalized (per picture)
time taken by analysts to perform quality checks with and without clues \ref{table:prod-2}. We
measure recall by checking the number
of annotations that were created from clues (similar to Table \ref{table:prod-1}) 

We found that a very high percentage of clues are being converted to actual annotations by the analysts. 
Table \ref{table:prod-1} shows five such sample jobs with an minimum of 95\% of all clues generated by the model being used by
the analysts without slowing down the overall process. 
All our analysts reported little disruption in their workflow. The average time to perform quality checks 
Human Quality Check 1 (QC1) and Human Quality Check 2 (QC2) 
between inspections with clues also improved, as demonstrated in Table \ref{table:prod-2}. More importantly, we also see
fewer missed damages when clues were used.

\begin{table}[ht] 
    \scriptsize
    \begin{center}
        \begin{tabular}{cccc}
            \toprule
            Job No. & No. of annotations & No. of clues converted & \% of clues converted \\
            \midrule
            1 & 183 & 178 & 97.3\% \\
            2 & 192 & 184 & 95.8\% \\
            3 & 124 & 124 & 100\% \\
            4 & 192 & 184 & 95.8\% \\
            5 & 192 & 184 & 95.8\% \\
            \bottomrule
        \end{tabular}
        \caption{Percentage of the clues generated by the model that were converted into actual annotations}
        \label{table:prod-1}
        \begin{tabular}{p{2cm}p{2cm}p{2cm}p{2cm}}
            \toprule
            Clues Used (yes/no) & Average QC1 minutes (per picture) & Average QC2 minutes (per picture) & Average number of missed damages (per inspection)\\
            \midrule
            no & 0.212 & 0.090 & 0.0080 \\
            yes & 0.205 & 0.086 & 0.0072 \\
            \bottomrule
        \end{tabular}
        \caption{Average time spent on QCs and number of misses in production}
        \label{table:prod-2}
    \end{center}
\end{table}
\vspace*{-6mm}

\section{Discussion and Future Work}
\label{discussion_and_future_work}

This work presented a system for automatically identifying damages on wind turbine blades,
which aids in lowering maintenance costs of wind farms. This is essential to enable wind energy's competitiveness with fossil fuels, 
the primary source of CO$_2$ emissions.
In order to achieve this goal, we define
a different set of metrics that are better proxies to our goal rather than the traditional
ML metrics. Additionally, we believe that the ultimate value of an ML system is driven
by the actual values it provides to end-users. Consistent with this thought process,
we started with a simple Mask R-CNN model for a simplified detection problem, presented the full pipeline
to users, and iteratively improved not just the model but also the manner in which it was utilized.
In short, we do not focus on the state-of-the-art results. Rather, we focus on the specific business use case.
This is especially important because even a few percentage improvement in time spent per inspection could save thousands of 
dollars at scale. Other factors are not as easily measured. If were were to scale up our operations, scaling our system 
is much easier than scaling the analyst team, which require additional hiring and training. All of these factors directly contribute to lowering the cost of
maintaining a wind farm and of wind energy at large.

There is a wide variety of promising future directions. While the system is built
around users' needs, we do recognize that there is room for improving the underlying detection model.
Our \textit{damage precision} was fairly low, which would result in extraneous suggestions.
One avenue to improve would be to train the model on the segmented blades. That is, the background will be segmented out of the images.
Once the model achieves certain performance, we would like to experiment with real-time online inference on the drone. This could aid the drone in recognizing and capturing regions of interest (damage areas) accurately.
Since the inspection process has no real-time human inputs, the model's baseline will need
to be much higher in this case.

While we have received encouraging results from our analysts during production tests, one of the notable suggestions from the team 
was to improve the shapes of the predicted boxes - more tightly drawn boxes around the damage. This is part of our future roadmap as well. The aim is to incorporate many ML products to enable automated windfarms \cite{iyer2021learning}.

\section{Acknowledgements}
Much of this work was made possible by the SkySpecs internal analyst team. We would also like to thank the anonymous reviewers for their insightful comments and feedback.

\setcitestyle{numbers}
\bibliography{neurips_2021}

\end{document}